\documentclass[lettersize,journal]{IEEEtran}
\usepackage{amsmath,amsfonts}
\pdfoutput=1
\usepackage{algorithmic}
\usepackage{algorithm}
\usepackage{array}
\usepackage[caption=false,font=normalsize,labelfont=sf,textfont=sf]{subfig}
\usepackage{textcomp}
\usepackage{stfloats}
\usepackage{url}
\usepackage{verbatim}
\usepackage{graphicx}
\usepackage{cite}
\usepackage{booktabs}
\usepackage{multirow}
\usepackage{bigstrut}
\usepackage{array}
\usepackage{tabularx}
\begin{document}

\title{Frequency-Domain Fusion Transformer for Image Inpainting}

\author{Sijin He,Guangfeng Lin*\thanks{*corresponding author},Tao Li,Yajun Chen}




\maketitle

\begin{abstract}
Image inpainting plays a vital role in restoring missing image regions and supporting high-level vision tasks, but traditional methods struggle with complex textures and large occlusions. Although Transformer-based approaches have demonstrated strong global modeling capabilities, they often fail to preserve high-frequency details due to the low-pass nature of self-attention and suffer from high computational costs. To address these challenges, this paper proposes a Transformer-based image inpainting method incorporating frequency-domain fusion. Specifically, an attention mechanism combining wavelet transform and Gabor filtering is introduced to enhance multi-scale structural modeling and detail preservation. Additionally, a learnable frequency-domain filter based on the fast Fourier transform is designed to replace the feedforward network, enabling adaptive noise suppression and detail retention. The model adopts a four-level encoder-decoder structure and is guided by a novel loss strategy to balance global semantics and fine details. Experimental results demonstrate that the proposed method effectively improves the quality of image inpainting by preserving more high-frequency information.
\end{abstract}


\section{Introduction}
Image inpainting has evolved into a critical task in image processing, requiring not only the restoration of missing regions but also a deep understanding of global structures to ensure the coherence, authenticity, and usability of the repaired image across various application domains. Compared to traditional methods, convolutional neural networks (CNN) offer significant advantages in image inpainting, such as automatic feature learning, improved texture synthesis, and better semantic understanding\cite{li2020recurrent,li2022misf,xie2019image}. However, CNN still struggle with capturing global context and long-range dependencies, which can lead to structural inconsistency and blurred results in complex or large missing regions.

In recent years, Transformer models have shown remarkable success in capturing global information through self-attention, overcoming the limitations of traditional CNN\cite{vaswani2017attention}. Initially achieving breakthroughs in machine translation, Transformers have been introduced into image inpainting tasks to leverage their global context modeling for reconstructing large missing regions\cite{zhou2021transfill,wan2021high,liang2021swinir,zhang2022swinfir,chen2023activating,chen2024hint,liu2025transref}. Their self-attention mechanism enables long-range dependency modeling, providing coherent and natural restoration results. Moreover, their modular design and high parallelism facilitate multi-scale feature fusion and cross-domain integration, especially in large-scale data environments.
\begin{figure}[htb]
	\begin{minipage}[b]{1.0\linewidth}
		\centering
		\centerline{\includegraphics[width=0.95\textwidth]{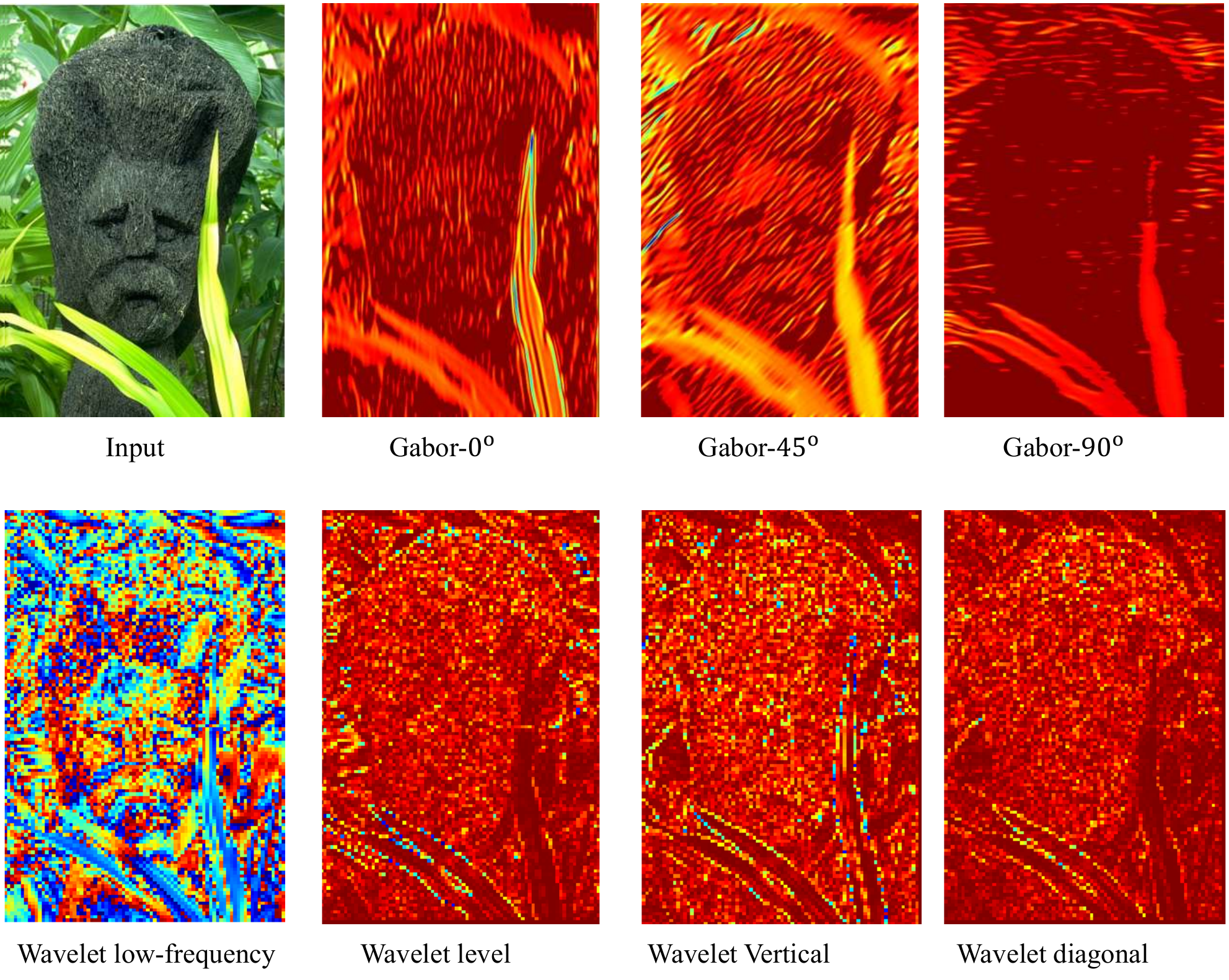}}
	\end{minipage}
	\caption{Comparison of Gabor and Wavelet Transform Effects on Multi-Directional Texture Image.Gabor filters provide strong directional selectivity for high-frequency textures, while wavelet transform excels at multiscale decomposition and global texture extraction.}
	\label{fig:res}
\end{figure}
While CNN excel at capturing local texture through neighboring pixel correlations, Transformers focus on global relationships by computing similarity between all positions in the input. However, in data-scarce scenarios, their ability to model local patterns weakens, affecting the restoration of fine textures and boundary consistency in image inpainting. Additionally, the quadratic complexity of self-attention makes Transformers computationally expensive for high-resolution images, limiting their scalability and efficiency. Standard Transformers also generate queries and keys from the same input without external guidance, which may restrict the effectiveness of attention. Designing more informative queries can enhance the model’s focus on critical regions and improve overall performance.

In this study, we propose a Transformer-based image inpainting method that incorporates frequency-domain fusion to enhance both global structure understanding and local texture reconstruction. The method combines wavelet transform and Gabor filtering to build a feature extraction mechanism with multi-scale and multi-directional perception capabilities. The wavelet transform enables effective decomposition of images into low and high frequency components across multiple scales, capturing both coarse structural layouts and fine-grained details\cite{torrence1998practical}. Building upon this, Gabor filters are applied to the high-frequency subbands to extract directional textures and edge features\cite{rai2020review}. To adapt to the characteristics of different subbands, we introduce a dynamic wavelength adjustment mechanism for Gabor filters, allowing the model to better capture texture details at various scales while avoiding underfitting or overfitting caused by fixed parameter settings. As shown in Fig. 1, Gabor filtering highlights directional high-frequency features more effectively, while wavelet transform captures global textures at multiple scales.Furthermore, we design a learnable frequency filter within the feedforward network to perform frequency-domain feature selection. By applying Fast Fourier Transform (FFT), the model dynamically suppresses redundant or noisy frequency components and retains only those features that are beneficial for image restoration. This approach improves reconstruction accuracy and efficiency, especially in scenarios with complex degradations.The overall model adopts a four-stage symmetric encoder-decoder architecture composed of multiple Transformer blocks. The encoder extracts global semantic information while progressively increasing channel capacity to enrich feature representations. The decoder then restores spatial resolution step-by-step, ensuring the final output maintains both semantic coherence and high-quality texture fidelity. A convolutional fusion layer is employed at the end to effectively integrate refined features with the original input.

The main contributions of our paper are summarized as follows:
\begin{itemize}
	\item We integrate wavelet-based multi-scale decomposition with adaptive Gabor filtering to jointly capture structural and textural information.
	\item We introduce a learnable frequency-domain filter using FFT to selectively retain meaningful features and suppress noise for more accurate restoration.
	\item We embed frequency-domain operations into the Transformer architecture to enhance its robustness and efficiency when handling complex image degradation.
\end{itemize}
\section{Related Work}
\subsection{Single Transformer Strategy}
In recent years, the Transformer model demonstrates outstanding capabilities in capturing global information through its self-attention mechanism, effectively overcoming the limitations of traditional convolutional neural networks. Initially successful in machine translation, Transformers also shorten training times by discarding recurrent and convolutional operations, encouraging researchers to apply this architecture to image inpainting tasks. Leveraging its global contextual understanding, the Transformer shows strong potential in reconstructing large missing regions.

Zhou et al.\cite{zhou2021transfill} are among the first to introduce Transformer architectures into image inpainting, achieving excellent results on images with extensive damage and complex depth information. However, their method struggles under low-light conditions or extreme lighting changes. To improve performance under such challenging scenarios, Wang et al.\cite{wang2022ft} propose a face inpainting approach that uses the Transformer to capture complex contextual relationships and locate missing regions. Their encoder-decoder framework progressively refines features to generate semantically coherent content, though it still shows limitations in detail restoration. Later, Zheng et al.\cite{zheng2022bridging} treat image inpainting as a sequence-to-sequence task, using the Transformer to model long-range dependencies and avoid interference from adjacent regions via a restricted CNN with small receptive fields. They also introduce an attention-aware layer to improve consistency between generated and visible regions. Additionally, Dong et al.\cite{dong2022incremental} develop an incremental Transformer-based model that modifies masked position encoding to handle diverse mask patterns. They also incorporate a structural feature extractor and a Fourier-based CNN texture recovery module to enhance the reconstruction of structure and texture in large missing regions.
\subsection{Multivariate Transformer Fusion}
Although Transformer models demonstrate powerful global modeling capabilities in image inpainting, they often struggle with restoring intricate local textures and structural details. To address this limitation, recent studies propose hybrid architectures that integrate Transformers with other models to enhance detail recovery while maintaining global context modeling. Liu et al.\cite{liu2022reduce} combine a variational autoencoder with a Transformer framework to construct a novel vector-quantized VAE model, enabling accurate reconstruction of missing regions while preserving undamaged content. They also introduce a Transformer variant without quantization, which enhances feature diversity prediction but incurs a heavy computational burden (~160GB), limiting its deployment in real-time applications. Li et al.\cite{li2022mat} propose a mask-aware Transformer incorporating multi-head contextual attention and dynamic masking, along with a style modulation module to improve generative diversity and computational efficiency. To further reduce complexity, Xu et al.\cite{ouyang2024image} develop Uformer-GAN, combining a Transformer with a GAN and applying post-processing refinement; however, its effectiveness is limited when initial restoration is poor. Zamir et al.\cite{zamir2022restormer} address high-resolution inpainting efficiency with Restormer, an optimized Transformer featuring long-range interaction and U-Net-style encoder-decoder balance. Building upon this, Phutke et al.\cite{phutke2023blind} introduce a wavelet-based multi-head attention mechanism that more effectively captures global and local features in blind inpainting tasks. In 2024, the HINT model\cite{chen2024hint} introduces a mask-prioritized downsampling module and a spatial-channel attention layer to enhance multiscale feature modeling while preserving visible information. Most recently, Jiang et al.\cite{jiang2024multi} propose a hybrid of Mamba and Transformer to overcome the trade-off between receptive field and efficiency. Mamba enables linear-complexity spatial modeling through selective scanning, while Transformer modules model channel dependencies. A multi-dimensional prompt learning module further enhances adaptability across various degradation types, though the method remains too slow for real-time applications.TransRef\cite{liu2025transref} proposes a transformer-based encoder-decoder network that progressively aligns and fuses reference features to effectively utilize reference images, addressing the challenge of image inpainting in complex semantic environments with diverse hole patterns.

\begin{figure*}[htb]
	\begin{minipage}[b]{1.0\linewidth}
		\centering
		\centerline{\includegraphics[width=1\textwidth,height=0.7\textwidth]{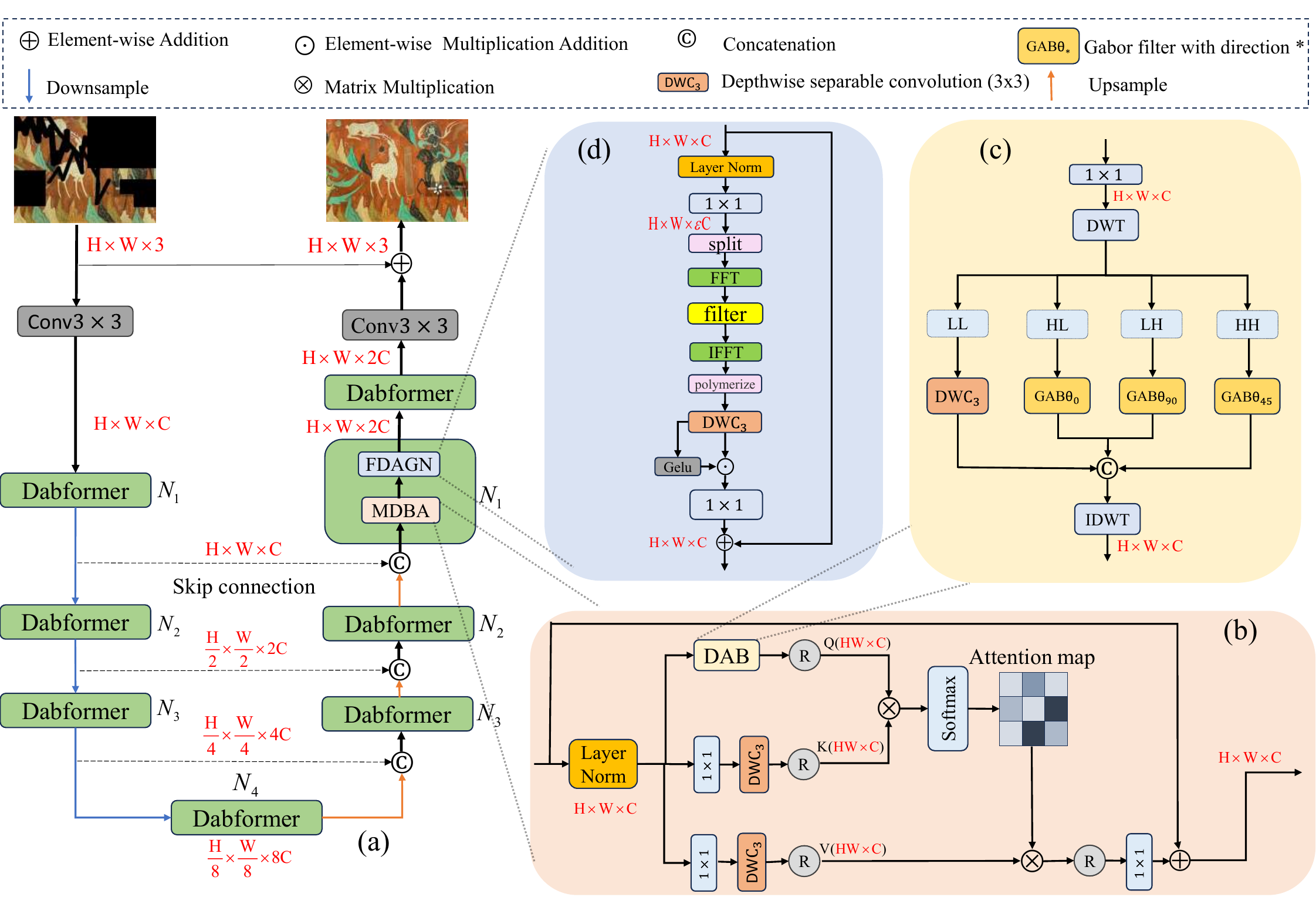}}
	\end{minipage}
	\caption{Detailed framework of Dabformer with the main constituent modules of 
		(a) Overall framework(Dabformer), (b) Frequency Domain Fusion Attention (FDFA), (c) Frequency Domain Fusion (FDF), (d) Frequency Domain Adaptive Gating Network (FDAGN)}
	\label{fig:res}
\end{figure*}
\section{Proposed Method}
\subsection{Overall Pipeline}
\label{ssec:subhead}
 Specifically, given a missing image ${I_{in}} \in {\mathbb{R}^{M\times 3}}$, it is first processed through a generalized convolutional layer along the RGB channel to form low-level feature embeddings ${X_t} \in {\mathbb{R}^{M \times C}}$, where ${M} = {{H \times W}}$, \textit{H} and \textit{W} are the height and width of the feature map \textit{M},  \textit{C} denotes the number of channels. Feature processing is performed  step-by-step by an encoder-decoder model with a 4-level symmetric structure, which each level consists of multiple transformer blocks. In order to extract feature representations layer by layer, spatial size reduction of the feature map and channel expansion are achieved by stepwise encoding and decoding operations in the adjacent layers of the encoder. Pixel-unshuffle and pixel-shuffle operations are used in the downsampling and upsampling process of the features\cite{shi2016real}. In addition, use skip connection to introduce more contextual information and facilitate the flow of information\cite{ronneberger2015u}. Finally, the refined features are processed through a convolutional layer and the resulting residual image is summed with the original input image to get the inpainting image, and the overall pipeline of our Dabformer architecture is shown in Fig 2. Formally, given the input feature ${X_{N - 1}}$ at the (N-1)-th block, the definition of the encoding process for Dabformer can be formulated as follows:
\begin{equation}\label{eq1}
	\begin{split}
	{X_N} = {{\tilde X}_N} + FDAGN(l({{\tilde X}_N})),\\
	{{\tilde X}_N} = {X_{N - 1}} + FDFA(l({X_{N - 1}})).
\end{split}
\end{equation}
where $l( \cdot )$ is the layer normalization, ${\tilde X_N}$ and ${ X_N}$ are output features of the FDFA and GTFFN modules.
\subsection{Frequency Domain Fusion Attention}
The Gabor filter is a spatial-frequency filter widely employed in the fields of image processing and computer vision, which uniquely combines Gaussian distribution and sinusoidal components to give it both spatial and frequency domain localization properties\cite{mehrotra1992gabor}. This filter responds to a variety of structures in an image at multiple scales and orientations, and exhibits excellent performance in texture and edge extraction, it is described as the follows:
\begin{gather}
	G(x,y;\lambda ,\theta ,\psi ,\sigma ,\gamma ) = \exp ( - \frac{{{{x'}^2} + {\gamma ^2}{{y'}^2}}}{{2{\sigma ^2}}})\cos (2\pi \frac{{x'}}{\lambda } + \psi ),\nonumber\\
	x' = x\cos \theta  + y\sin \theta ,\nonumber\\
	y' =  - x\sin \theta  + y\cos \theta {\rm{.}}
\end{gather}
where \textit{x} and \textit{y} are the horizontal and vertical pixel coordinates , $\lambda$ denotes the filter wavelength, $\theta$ is the main direction of the filter, $\psi$ denotes the phase shift, $\sigma$ is the standard deviation of the Gaussian distribution, $\gamma$ denotes the spatial ellipticity of the filter, ${x'}$ and ${y'}$ are the coordinates obtained by rotating \textit{x} and \textit{y} in the original coordinate system. The larger the wavelength, the wider the range of image structures perceived by the filter.

Gabor filter are effective for enhancing local textures and capturing directional features. However, their fixed-scale nature limits their adaptability to features of varying sizes, making it difficult to balance fine details and large structures in complex image inpainting tasks. This limitation reduces their ability to provide sufficient context, especially when dealing with large missing regions and intricate textures.Wavelet transform, on the other hand, offers inherent multi-scale analysis by decomposing an image into sub-bands (LL, HL, LH, HH), enabling the model to capture both global structure and fine details. This makes it more suitable for tasks requiring cross-scale understanding.Nevertheless, wavelet transform lacks strong directionality. While it performs well in scale decomposition, it cannot explicitly extract orientation-specific textures like Gabor filters. This limits its effectiveness when dealing with images containing complex, directional patterns. Therefore, combining both methods can help leverage their complementary strengths.Based on this, we propose a feature extraction mechanism that combines wavelet transform and Gabor filter to achieve both multi-scale and multi-directional perception. The wavelet transform effectively separates low- and high-frequency components at different scales, capturing global structures and fine details. Gabor filter are then applied to the wavelet-derived high-frequency features to extract directional textures and edges. The wavelength of the Gabor filter is dynamically adjusted based on sub-band characteristics, reducing the risk of underfitting or overfitting due to fixed parameters. 
\subsubsection{Frequency Domain Fusion}
We decompose the feature map into four sub-bands via discrete wavelet transform: three high-frequency sub-bands (HL, LH, HH) and one low-frequency sub-band (LL). Each high-frequency component captures directional textures and edges, and is enhanced using Gabor filters aligned to its dominant orientation—horizontal for HL, vertical for LH, and diagonal for HH—thereby refining texture representation in different directions.For the LL sub-band, which contains the global structure and smooth region information of the image, we apply a 3×3 depthwise separable convolution to capture spatial context while preserving structural integrity and minimizing artifacts. Unlike high-frequency components, LL is less sensitive to directionality and benefits more from efficient spatial modeling.

To further improve flexibility, an adaptive wavelength selection strategy is introduced. By adjusting the wavelength of Gabor filters based on local image features and context, the model can better respond to texture variations at different scales and orientations.Instead of relying on fixed or heuristically set wavelengths, we treat the wavelength as a learnable parameter initialized to a reasonable prior value. During training, this parameter is optimized jointly with the rest of the network using gradient backpropagation, allowing the model to automatically adjust filter scales based on the task and data.This adaptive fusion of wavelet-based multi-scale structure and Gabor-based directional texture enhances the model’s capability to reconstruct fine details and maintain global consistency, especially in cases of complex degradation and large missing regions.
\subsubsection{Attention Module}
Inspired by Restormer, We use cross-channel attention instead of traditional self-attention to save computational complexity. Given a layer normalized input tensor  ${X_{N - 1}} \in {\mathbb{R}^{M \times C}}$,   which is given as:
\begin{gather}
Q = {f_{1 \times 1}}(FDF(X_{N - 1}))
\end{gather}
where $FDF( \cdot )$ is the frequency domain fusion operation, ${f_{1 \times 1}}( \cdot )$ is a 1×1 point-by-point convolution.Calculating Attention can be expressed as follows:
\begin{equation}\label{eq1}
	\begin{split}
	Att(Q,K,V) = \varphi (Q,K)V,\\
	{\rm{ }}\varphi (Q,K) = softmax (\frac{{Q}{K^T}}{\xi }),
\end{split}
\end{equation}
where ${\xi }$  is a scale parameter that can be learned to control the size of the attention map, K is the key vector and V is the value vector.. By performing the attentional computation in the channel dimension, we change the computational complexity from $o(M \times M)$ to $o(C \times C)$ ,which is reduced due to $M \gg C$. Similarly, we divide the channels into multiple heads and train attention mappings for each head independently in parallel to improve the model's ability to capture complex relationships. Eventually, the output ${{\tilde X}_N}$ of the attention can be shown as:
\begin{gather}
	{{\tilde X}_N} = {f_{1 \times 1}}(Att(Q,K,V)) + X_{N - 1}
\end{gather}
\subsection{Frequency Domain Adaptive Gating Network}
In attention mechanisms, wavelet transforms are effective at decomposing images into multiple scales and frequency bands, thereby capturing multi-level information across various frequency ranges. This is particularly useful for extracting local textures and fine-grained details. However, wavelet transforms exhibit weak directional selectivity, making them less capable of handling complex texture orientations. In the three high-frequency sub-bands after decomposition, some features may be redundantly represented—for instance, certain textures may appear simultaneously in both horizontal and diagonal sub-bands—resulting in redundancy or suboptimal feature assignment, which reduces information utilization efficiency.
In contrast, Gabor filters provide strong directional selectivity and are particularly effective for texture-rich images with distinct edges. However, since the directional filters in Gabor transforms are predefined and operate locally—only considering pixel neighborhoods—real-world texture orientations that fall between preset angles may not be effectively enhanced, leading to potential information redundancy.

To address the redundancy and loss issues inherent in wavelet and Gabor-based feature extraction, we propose a Frequency-Domain Adaptive Gating Network (FDAGN). This module is designed to leverage the advantages of frequency-domain representation for more precise selection and enhancement of salient information, thereby improving the representational capacity and restoration performance of the model.

The input features are first normalized using layer normalization to improve training stability. Then, a 1×1 convolution is applied to expand the feature channels, enhancing the expressiveness of the feature space. The resulting feature maps are divided into multiple sub-blocks, and each sub-block undergoes a Fourier transform to project spatial-domain features into the frequency domain, allowing for explicit separation of different frequency components.

In the frequency domain, we design a learnable filter composed of complex convolutions to process the complex-valued features obtained via Fourier transform. The filter is initialized close to an identity mapping to preserve the original frequency information as much as possible. During training, the convolutional kernels are optimized through backpropagation, enabling adaptive modeling across the frequency, channel, and spatial dimensions. This mechanism effectively suppresses redundant and noisy components in the frequency domain, highlights structurally and texturally salient features, and enhances the model’s ability to perceive meaningful information under complex backgrounds.

After filtering, the features are transformed back into the spatial domain using an inverse Fourier transform. This ensures that subsequent processing occurs at the pixel level, preserving structural integrity and fine details. The following operations remain consistent with our method in Gabformer\cite{10893764}: the transformed features are processed with a 3×3 depthwise separable convolution to further enhance representation. Finally, the feature maps are split into two paths, with one path applying the GELU activation function to perform additional spatial-domain feature refinement.
\subsection{Loss function}
The L1 loss demonstrates strong convergence and stability in image inpainting tasks, effectively measuring the pixel-level differences between the restored image and the ground truth. It guides the model to produce results that are numerically close to the original image. However, since L1 loss focuses solely on absolute pixel errors, it struggles to capture high-level semantic information, multi-scale texture details, and structural consistency. As a result, the restored images may still suffer from blurred details, missing textures, or incomplete structures in terms of visual quality.

To address these limitations and enhance the perceptual quality and visual coherence of the restored images, we further introduce perceptual loss, edge loss, and structural similarity loss based on the L1 loss. These additional loss components provide multi-level constraints from the feature, edge, and structural perspectives, thereby improving detail reconstruction and enhancing the overall visual performance.The perceptual loss is defined as the difference between the deep feature maps of the restored image and the ground truth image extracted by a pre-trained VGG16 model \cite{simonyan2014very} as:
\begin{gather}
{L_P} = \sum\limits_{s = 1}^S {(||{\phi _s}({G_t}) - {\phi _s}(O)|{|_1}} )
\end{gather}
where ${{G_t}}$ denotes the ground truth image, $O$ denotes the output image, and ${\phi _s}$ represents the feature map from the s-th layer of the pre-trained VGG16 model, $s \in (1,S)$. To enhance edge information during training edge loss is introduced and computed as the difference between edges extracted by the Sobel operator given as:
\begin{gather}
{L_E} = ||Sob({G_t}) - Sob(O)|{|_1}
\end{gather}
where $Sob( \cdot )$ denotes the Sobel operator. To generate structurally consistent restoration results structural similarity loss is introduced and defined as:
\begin{gather}
{L_M} = 1 - SSIM(O)
\end{gather}
where $SSIM(\cdot )$ is the structural similarity calculation. The total loss function of the model is defined as:
\begin{gather}
{L_T} = {\lambda _1}{L_1} + {\lambda _P}{L_P} + {\lambda _E}{L_E} + {\lambda _M}{L_M}
\end{gather}
where ${L_1}$ denotes the L1 loss, ${\lambda_1}$ is the weight of the L1 loss, ${\lambda_P}$ is the weight of the perceptual loss, ${\lambda_E}$ is the weight of the edge loss, and ${\lambda_M}$ is the weight of the structural similarity loss. These weights are empirically set as: $\lambda_1 = 10$, $\lambda_P = 0.6$, $\lambda_E = 0.4$, and $\lambda_M = 0.5$.
\begin{table*}[htbp]
	\centering
	\caption{Quantitative Results of Different Methods on Image Deraining. \textbf{Bold} indicates best results.}
	\normalsize
	\renewcommand{\arraystretch}{1.3}
	\tabcolsep=0.27cm
	\begin{tabular}{
			>{\centering\arraybackslash}p{7em} 
			>{\centering\arraybackslash}p{6em} 
			>{\centering\arraybackslash}p{6em} 
			>{\centering\arraybackslash}p{6em} 
			>{\centering\arraybackslash}p{6em} 
			>{\centering\arraybackslash}p{5.5em}}
		\toprule
		\multirow{2}{*}{Methods} & \multicolumn{4}{c}{Datasets} & \multirow{2}{*}{Parameter} \\
		\cmidrule(lr){2-5}
		& Rain200L & Rain200H & DDN-Data & DID-Data & \\
		\midrule
		DSC\cite{luo2015removing}        & 27.16/0.866 & 14.73/0.381 & 27.31/0.837 & 24.24/0.828 & / \\
		GMM\cite{li2016rain}        & 28.66/0.865 & 14.50/0.416 & 27.55/0.848 & 25.81/0.834 & / \\
		MSPFN\cite{jiang2020multi}      & 38.58/0.983 & 29.36/0.903 & 32.99/0.933 & 33.72/0.955 & 13.35M \\
		PReNet\cite{ren2019progressive}     & 37.80/0.981 & 29.04/0.899 & 32.60/0.946 & 33.17/0.948 & 0.17M \\
		MPRNet\cite{zamir2021multi}     & 39.47/0.982 & 30.67/0.911 & 33.10/0.935 & 33.99/0.959 & 3.63M \\
		SwinIR\cite{liang2021swinir}     & 40.61/0.987 & 31.76/0.915 & 33.16/0.931 & 34.07/0.931 & 15.03M \\
		Restormer\cite{zamir2022restormer}  & 40.99/0.989 & 32.00/0.932 & 34.20/0.957 & 35.29/0.964 & 26.12M \\
		IDT\cite{xiao2022image}        & 40.74/0.988 & 32.10/0.934 & 33.84/0.955 & 34.89/0.962 & 115.50M \\
		HCT-FFN\cite{chen2023hybrid}    & 39.70/0.985 & 31.51/0.910 & 33.00/0.950 & 33.96/0.959 & 0.87M \\
		DRSformer\cite{chen2023learning}  & 41.23/0.989 & 32.18/0.933 & 34.36/0.959 & 35.38/0.965 & 33.65M \\
		FADformer\cite{gao2024efficient}  & \textbf{41.69/0.990} & 32.30/0.936 & \textbf{34.42/0.960} & 35.48/0.966 & 22.89M \\
		Dabformer  & 41.66/0.990 & \textbf{32.34/0.936} & 34.09/0.957 & \textbf{35.53/0.966} & 29.73M \\
		\bottomrule
	\end{tabular}
	\label{tab:rain-results}
\end{table*}
\begin{figure*}[htb]
	\begin{minipage}[b]{1.0\linewidth}
		\centering
		\centerline{\includegraphics[width=0.95\textwidth]{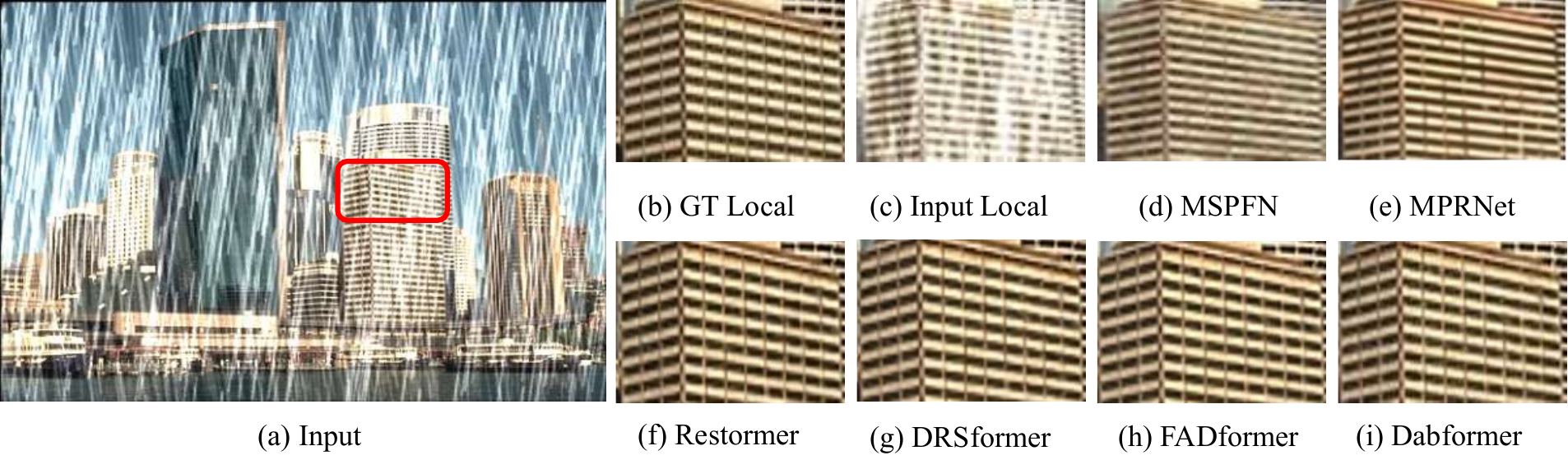}}
	\end{minipage}
	\caption{Qualitative comparison of deraining methods on the Rain200H dataset. From left to right: (a) Input rainy image, (b) Ground Truth (GT) local region, (c) Input local, (d) MSPFN, (e) MPRNet, (f) Restormer, (g) DRSformer, (h) FADformer, and (i) Dabformer. The zoomed-in regions highlight our method’s ability to restore fine details while maintaining natural image appearance.}
	\label{fig:res}
\end{figure*}
\begin{figure*}[htb]
	\begin{minipage}[b]{1.0\linewidth}
		\centering
		\centerline{\includegraphics[width=0.95\textwidth]{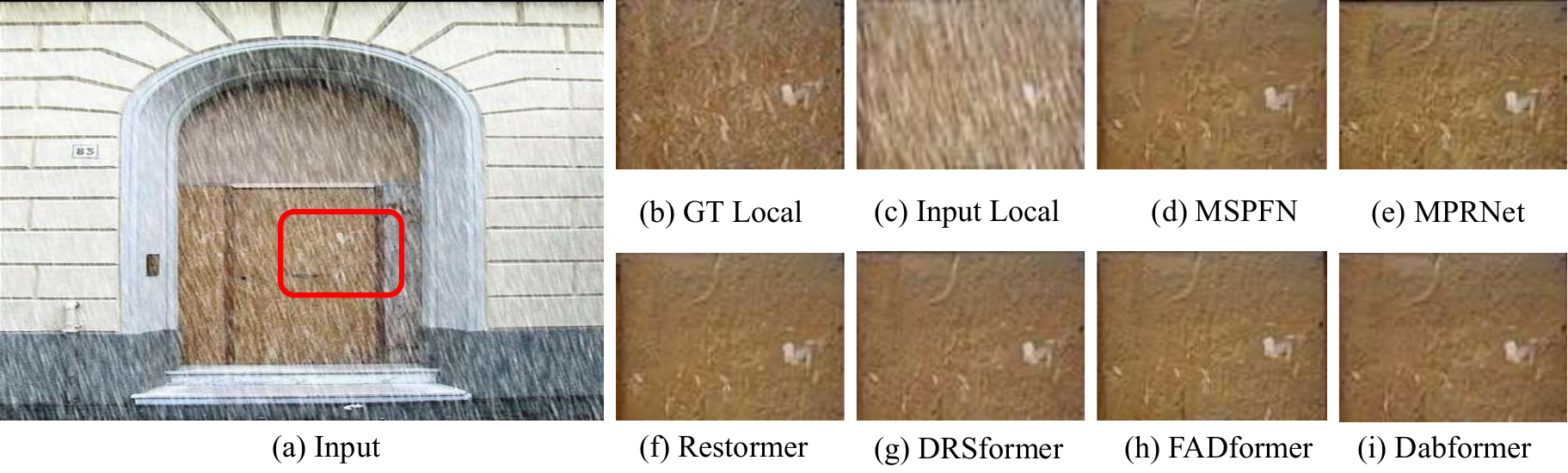}}
	\end{minipage}
	\caption{Qualitative comparison of deraining methods on the DID-Data dataset. From left to right: (a) Input rainy image, (b) Ground Truth (GT) local region, (c) Input local, (d) MSPFN, (e) MPRNet, (f) Restormer, (g) DRSformer, (h) FADformer, and (i) Dabformer. The zoomed-in regions highlight our method’s ability to restore fine details while maintaining natural image appearance.}
	\label{fig:res}
\end{figure*}
\begin{figure*}[htb]
	\begin{minipage}[b]{1.0\linewidth}
		\centering
		\centerline{\includegraphics[width=0.95\textwidth]{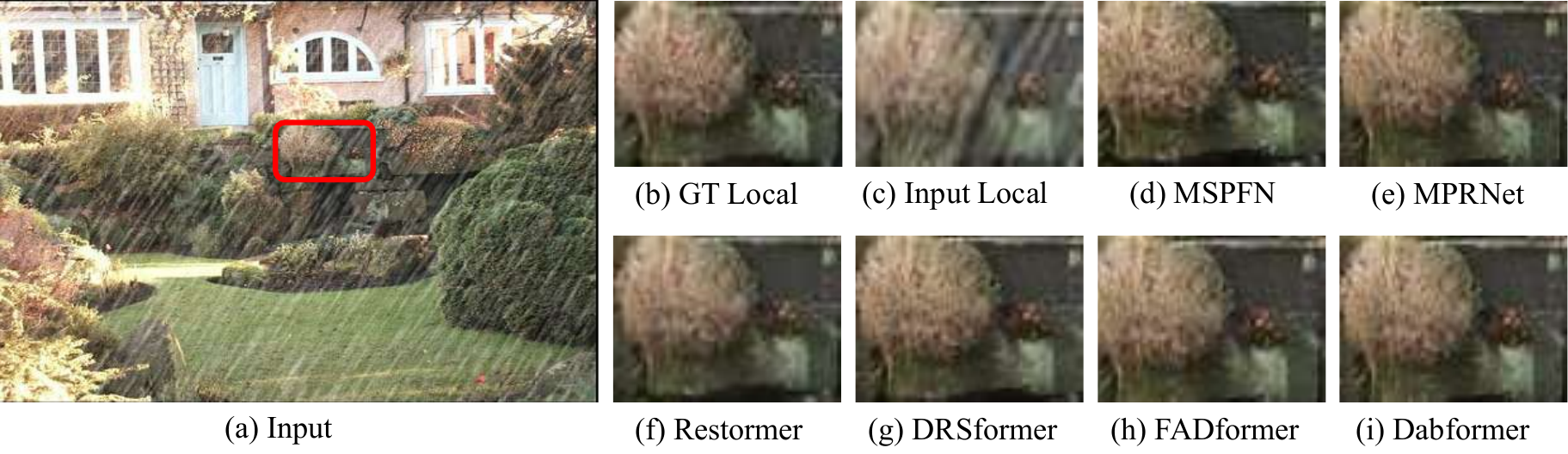}}
	\end{minipage}
	\caption{Qualitative comparison of deraining methods on the DDN-Data dataset. From left to right: (a) Input rainy image, (b) Ground Truth (GT) local region, (c) Input local, (d) MSPFN, (e) MPRNet, (f) Restormer, (g) DRSformer, (h) FADformer, and (i) Dabformer. The zoomed-in regions highlight our method’s ability to restore fine details while maintaining natural image appearance.}
	\label{fig:res}
\end{figure*}
\section{EXPERIMENT}
\subsection{Image Rain Removal Dataset}
In this work the adaptability of the proposed model is demonstrated by starting from the relatively simple task of image deraining and gradually extending to the more complex task of degraded image restoration. The datasets used for training and testing in image deraining are as follows. Rain200H/L\cite{yang2017deep} contains rain streaks in five different directions with 1800 image pairs in the training set and 200 image pairs in the test set, totaling 400 images. Rain200H features denser rain streaks compared to Rain200L. DDN-Data\cite{fu2017removing} consists of 12600 training image pairs and 1400 testing image pairs. Each clean image can generate 14 synthesized images with different rain directions and densities. DID-Data\cite{zhang2018density} includes 12000 training image pairs and 1200 testing image pairs, covering three levels of rain density: light, medium, and heavy. Each pair contains a clean image and a corresponding rainy image.
\subsection{Damaged Image Restoration Dataset}
For the degraded image restoration task the datasets used are Places2\cite{zhou2017places} and CelebA\cite{liu2015deep}. Places2 contains approximately 10 million images, with 5000 to 30000 training samples per category. Due to computational resource constraints, 200 categories were randomly selected with 600 images each, resulting in a total of 120000 images for training. An additional 6000 images were used for testing. CelebA is a large-scale high-resolution face dataset containing 30000 images, with 26000 used for training and 4000 for testing. To improve robustness, each image in the training and testing sets was randomly corrupted by noise blocks of varying size and position, simulating real-world image degradation such as occlusion, damage, or interference.
\subsection{Implementation Details}
\subsubsection{Model Details}
In this work, the proposed model adopts a four-layer encoder-decoder architecture. From the first to the fourth layer the number of Transformer blocks is set to 4, 6, 6, and 8 respectively, enabling the extraction of increasingly deeper features in the later stages. The number of attention heads in each layer is set to 1, 2, 4, and 8 accordingly. For the fixed parameters of the Gabor filters the standard deviation $\sigma $ is set to $2\pi $, the phase offset $\psi $ is set to 0, and the spatial aspect ratio $\gamma $ is set to 0.5.
\subsubsection{Training Details}
The initial learning rate is set to $2 \times {10^{ - 4}}$ and gradually reduced to $1 \times {10^{ - 6}}$ using the cosine annealing strategy . The model is trained for 1400000 iterations with a channel expansion ratio of 2.66. The AdamW optimizer is used to regularize the model. All experiments in this paper were conducted on the same device, i.e., a single NVIDIA RTX 3090Ti GPU and  a 3.40 GHz AMD Ryzen 5700X CPU.
\subsection{Evaluation Metrics}
To comprehensively evaluate the performance of image inpainting at both the pixel level and the structural level, ensuring that the results are both objective and aligned with human visual perception, this paper adopts Peak Signal-to-Noise Ratio (PSNR) and Structural Similarity Index Measure (SSIM) as quantitative evaluation metrics.
\subsection{Experiments on Image Deraining}
\subsubsection{ Comparative Methods}
In the image deraining task, we compare our proposed method with several state-of-the-art approaches. These include two prior-based models, two CNN-based methods, and several Transformer-based models. The CNN-based methods include MSPFN\cite{jiang2020multi}, PReNet\cite{ren2019progressive}, and MPRNet\cite{zamir2021multi}, while the Transformer-based methods include SwinIR\cite{liang2021swinir}, Restormer\cite{zamir2022restormer}, HCT-FFN\cite{chen2023hybrid}, IDT\cite{xiao2022image}, DRSformer\cite{chen2023learning}, and FADformer\cite{gao2024efficient}. The best  result is highlighted in bold. “/” indicates unavailable data.
\subsubsection{ Quantitative Results}
The table I presents the comparison results of different image deraining methods across multiple public datasets. From the results, it is evident that the performance of various deraining approaches varies significantly depending on the dataset. Our proposed Dabformer demonstrates excellent performance on datasets with dense and complex rain streaks such as Rain200H and DID-Data. Notably, on Rain200H, it achieves a PSNR of 32.34 dB and an SSIM of 0.936, significantly outperforming mainstream methods. This indicates that our model possesses strong modeling capability and detail restoration ability in handling complex, multi-directional, and multi-scale rain streaks.

In contrast, on datasets with sparse rain streaks such as Rain200L and DDN-Data, although our method still maintains a leading overall performance, it shows a slight decrease compared to FADformer. This can be attributed to the nature of dense rain streaks which exhibit clear directionality and continuity, making Gabor filters effective in capturing these features. Combined with wavelet transforms that extract multi-scale high-frequency information, this enables the model to precisely focus on rain regions and improve removal results. However, sparse rain streaks are scattered and lack fixed directions, causing their high-frequency features to be easily confused with background details, thereby reducing the model’s discrimination ability and affecting deraining performance.

Furthermore, by introducing a high-frequency texture guidance mechanism alongside an adaptive gating network in the frequency domain, our method achieves stable and outstanding results across multiple datasets while effectively controlling parameter count. This demonstrates its efficiency and practical applicability.
\subsubsection{ Qualitative Results}
The qualitative results of various methods on the image deraining task are illustrated in Fig.3-Fig.5. These comparisons visually demonstrate the effectiveness of each method in removing rain streaks and restoring image details. The proposed Dabformer not only effectively eliminates rain streaks but also significantly improves the recovery of structural and textural details. Compared with existing mainstream approaches, our method achieves notable advantages in image quality, detail preservation, and structural consistency.

In particular, Dabformer demonstrates a strong ability to balance rain removal and texture retention when dealing with complex rain patterns and fine-grained details. On datasets such as Rain200H and DID-Data, the model successfully removes intensive and directionally consistent rain streaks while restoring intricate local textures and maintaining coherent global structures. On the DDN-Data dataset, where rain streaks are sparse and randomly distributed, the model faces greater challenges in distinguishing between rain artifacts and background details. However, even under such challenging conditions, Dabformer still delivers visually consistent and high-quality results. Moreover, despite using fewer parameters, Dabformer consistently maintains strong visual performance and restoration accuracy across diverse datasets, validating its adaptability and robustness in complex degradation scenarios.
\begin{table*}[htbp]
	\centering
	\caption{Quantitative Results of Different Methods on Damaged Image Restoration. \textbf{Bold} indicates best results.}
	\renewcommand{\arraystretch}{1.3}
	\tabcolsep=0.27cm 
	\normalsize
	\begin{tabular}{
			>{\centering\arraybackslash}p{6.445em} 
			>{\centering\arraybackslash}p{4.61em} 
			>{\centering\arraybackslash}p{4.61em} 
			>{\centering\arraybackslash}p{4.72em} 
			>{\centering\arraybackslash}p{4.61em} 
			>{\centering\arraybackslash}p{4.61em} 
			>{\centering\arraybackslash}p{4.78em} 
			>{\centering\arraybackslash}p{3.945em}}
		\hline
		\multirow{3}[6]{*}{Methods} & \multicolumn{6}{c}{Datasets} & \multirow{3}[6]{*}{Parameter} \\
		\cline{2-7}
		& \multicolumn{3}{c}{Places2} & \multicolumn{3}{c}{CelebA} & \\
		\cline{2-7}
		& 20\%-30\% & 40\%-50\% & 60\%-70\% & 20\%-30\% & 40\%-50\% & 60\%-70\% & \\
		\hline
		RFR\cite{li2020recurrent}              & 24.92/0.850 & 21.14/0.717 & 18.30/0.697 & 25.39/0.921 & 20.03/0.854 & 17.93/0.652 & 12.66M \\
		ICT\cite{wan2021high}              & 25.09/0.901 & 21.28/0.842 & 18.49/0.708 & 26.40/0.939 & 21.84/0.877 & 19.04/0.729 & 29.73M \\
		CMT\cite{guo2022cmt}              & 25.80/0.923 & 21.53/0.857 & 19.86/0.761 & 28.24/0.952 & 23.78/0.900 & 21.04/0.806 & 15.21M \\
		Restormer\cite{zamir2022restormer}        & 25.93/0.929 & 21.61/0.824 & 19.61/0.774 & 28.07/0.945 & 22.10/0.882 & 19.83/0.776 & 26.12M \\
		ZITS\cite{dong2022incremental}             & 25.56/0.902 & 22.03/0.771 & 20.03/0.781 & 28.03/0.947 & 23.45/0.891 & 20.86/0.803 & 31.48M \\
		RePaint*\cite{lugmayr2022repaint}        & 25.97/0.930 & 21.99/0.852 & 20.10/0.803 & 29.01/0.969 & 23.92/0.912 & 21.50/0.811 & 103.41M \\
		IR-SDE\cite{luo2023image}           & 25.52/0.928 & 21.53/0.867 & 19.53/0.813 & 28.85/0.966 & 23.76/0.910 & 20.93/0.806 & 92.00M \\
		StrDiffusion\cite{liu2024structure}      & \textbf{26.30/0.950} & \textbf{22.39/0.882} & \textbf{20.43/0.858} & \textbf{29.31/0.971} & 24.50/0.923 & \textbf{21.75/0.874} & 114.05M \\
		Dabformer & 26.18/0.938 & 22.42/0.780 & 20.04/0.792 & 29.05/0.967 & \textbf{26.62/0.945} & 21.44/0.842 & 29.73M \\
		\hline
	\end{tabular}
	\label{tab:addlabel}
\end{table*}
\begin{figure*}[htb]
	\begin{minipage}[b]{1.0\linewidth}
		\centering
		\centerline{\includegraphics[width=0.95\textwidth]{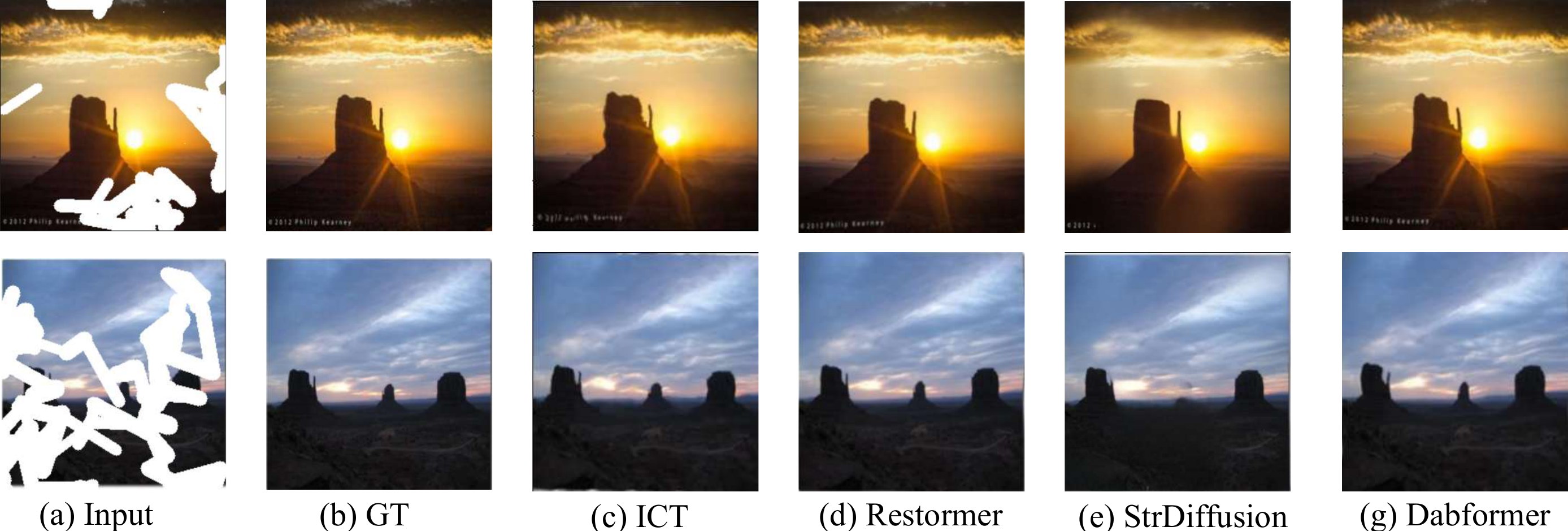}}
	\end{minipage}
	\caption{Qualitative comparison of image inpainting methods on the Places2 dataset with different mask ratios. From left to right: (a) Input, (b) Ground Truth, (c) ICT, (d) Restormer, (e) StrDiffusion, and (f) Dabformer.}

	\label{fig:res}
\end{figure*}
\begin{figure*}[htb]
	\begin{minipage}[b]{1.0\linewidth}
		\centering
		\centerline{\includegraphics[width=0.95\textwidth]{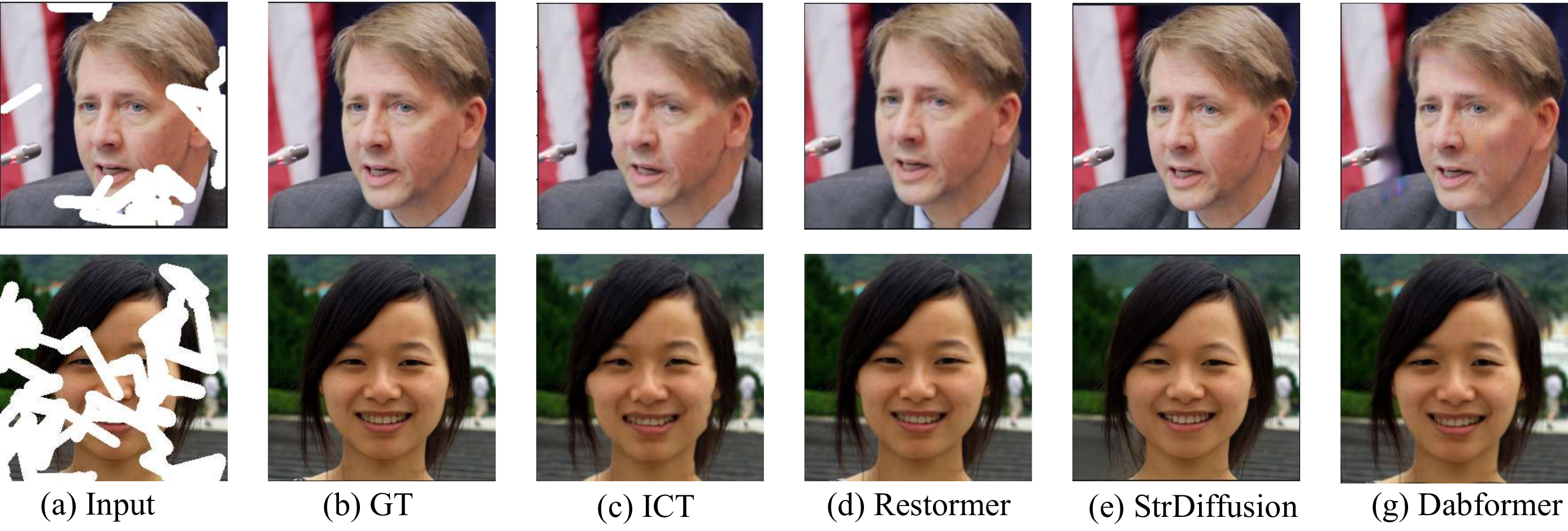}}
	\end{minipage}
		\caption{Qualitative comparison of image inpainting methods on the CelebA dataset with different mask ratios. From left to right: (a) Input, (b) Ground Truth, (c) ICT, (d) Restormer, (e) StrDiffusion, and (f) Dabformer.}
	\label{fig:res}
\end{figure*}
\subsection{Experiments on Damaged Image Restoration}
\subsubsection{ Comparative Methods}
In the task of restoring corrupted images, we compare our proposed method with several state-of-the-art approaches. These include CNN-based methods (such as RFR\cite{li2020recurrent}), Transformer-based methods that leverage global dependencies (such as ICT\cite{wan2021high}, CMT\cite{guo2022cmt}, and Restormer\cite{zamir2022restormer}), structure-guided methods that neglect semantic consistency (such as ZITS\cite{dong2022incremental}), and diffusion-based models (such as RePaint\cite{lugmayr2022repaint}, IR-SDE\cite{luo2023image}, and StrDiffusion\cite{liu2024structure}).To ensure a fair comparison, we evaluate RePaint using the IR-SDE pretrained model, denoted as RePaint*.
\subsubsection{ Quantitative Results}
The quantitative results of our comparisons for corrupted image restoration are presented in Table II. The proposed Dabformer achieves consistently superior performance compared to representative state-of-the-art methods across varying occlusion levels. On the Places2 dataset, it achieves a PSNR of 22.42 under the {\rm{40\%}} to {\rm{50\%}} occlusion range, outperforming StrDiffusion and Restormer. Under the highest occlusion level of {\rm{60\%}} to {\rm{70\%}}, Dabformer still maintains an advantage in both PSNR and SSIM, demonstrating its robustness in severely corrupted scenarios.

Similarly, on the CelebA dataset, Dabformer achieves the highest PSNR and SSIM scores under both moderate and heavy occlusion levels. Notably, it surpasses leading methods such as Restormer and ZITS under the {\rm{40\%}} to {\rm{50\%}} and {\rm{60\%}} to {\rm{70\%}} occlusion settings. These results validate the effectiveness of the proposed model in restoring detailed structures and textures in complex scenes.
\subsubsection{ Qualitative Results}
Qualitative results of various methods for the damaged image restoration task, as shown in Fig. 6–Fig. 7, demonstrate that Dabformer achieves more coherent and visually pleasing reconstructions compared to existing state-of-the-art approaches. In natural scene images, Dabformer effectively preserves color consistency and structural integrity across large regions, mitigating common artifacts such as color blocks, texture discontinuities, and geometric distortions often observed in other methods.

For face image restoration, Dabformer generates more natural facial features with enhanced completeness and realism, particularly in key regions such as the eyes, nose, and mouth. It also suppresses visible artifacts, resulting in smoother transitions and more realistic outputs. In terms of detail restoration, Dabformer demonstrates a strong ability to recover fine textures and maintain global structural consistency. Although it still lags slightly behind methods like StrDiffusion in rendering ultra-fine textures and achieving photorealism, the overall restoration quality is competitive, with clear potential for future improvement in high-fidelity detail generation.
\subsection{Ablation Study}
\subsubsection{Component ablation experiment}
To validate the effectiveness of the proposed multi-head attention module that integrates wavelet and Gabor transforms, as well as the Fourier-based gated frequency filtering network for feature selection, a series of ablation experiments were conducted on the CelebA dataset with an occlusion range of  {\rm{40\%}} to  {\rm{50\%}}. Specifically, $Q_D$ represents the use of 2D discrete wavelet transform to extract features from the query vector, $Q_G$ represents the use of Gabor filters to extract features, and $Q_{D+G}$ represents the use of the proposed frequency-domain fusion processing module to extract features. FDAGN refers to the frequency-domain adaptive gating network proposed in this paper. This paper first compares the effects of different feature extraction methods on model performance, including using only wavelet transform for multi-scale decomposition, using only Gabor transform for extracting directional texture features, and the combination of wavelet and Gabor features. Furthermore, to evaluate the effectiveness of the gating-based feature selection module, this paper compares the results with and without this module.
The quantitative results are reported in Table III, and the corresponding qualitative results are illustrated in Fig. 8.

The study first analyzes the impact of different feature extraction strategies on model performance, including using discrete wavelet transform alone for multi-scale decomposition, using Gabor filters alone for directional texture extraction, and combining both methods through a frequency-domain fusion mechanism. The results indicate that wavelet transform alone enhances the model’s ability to capture multi-scale contextual information, while Gabor filters significantly improve structural detail reconstruction by providing directional cues.

The joint use of wavelet and Gabor features achieves the best performance, demonstrating their complementary strengths and effectiveness in restoring complex textures and global structures. Additionally, experiments show that removing the frequency-domain gated filtering module leads to a performance drop. This suggests that conventional feedforward networks lack the capability to selectively enhance relevant features and suppress redundant ones, thereby limiting reconstruction quality. In contrast, the proposed gated network leverages Fourier transform to operate in the frequency domain, enabling global feature enhancement and adaptive selection through gating. This contributes to superior detail restoration and confirms the module’s effectiveness in targeted feature refinement.

\begin{table}[htbp]
	\centering
	\caption{Ablation study with quantitative results on the CelebA dataset under {\rm{40\%}} to  {\rm{50\%}} occlusion ratio for image inpainting with different network configurations}
	\renewcommand{\arraystretch}{1.3} 
	\setlength{\tabcolsep}{12pt} 
	\begin{tabularx}{\linewidth}{c *{2}{>{\centering\arraybackslash}X}}
		\hline
		\multirow{2}{*}{Network Configuration} & \multicolumn{2}{c}{Evaluation} \\
		\cline{2-3}
		& PSNR & SSIM \\
		\hline
		$\mathit{Q}, \mathit{K}, \mathit{V} + \mathit{FFN}$ & 21.95 & 0.8844 \\
		$\mathit{Q_D}, \mathit{K}, \mathit{V} + \mathit{FFN}$ & 23.80 & 0.9107 \\
		$\mathit{Q_G}, \mathit{K}, \mathit{V} + \mathit{FFN}$ & 22.96 & 0.8991 \\
		$\mathit{Q_{D+G}}, \mathit{K}, \mathit{V} + \mathit{FFN}$ & 24.71 & 0.9246 \\
		$\mathit{Q_{D+G}}, \mathit{K}, \mathit{V} + \mathit{FDAGN}$ & \textbf{26.62} & \textbf{0.9452} \\
    \hline
\end{tabularx}
\label{tab:addlabel}
\end{table}
\begin{figure}[htb]
	\begin{minipage}[b]{1.0\linewidth}
		\centering
		\centerline{\includegraphics[width=0.95\textwidth]{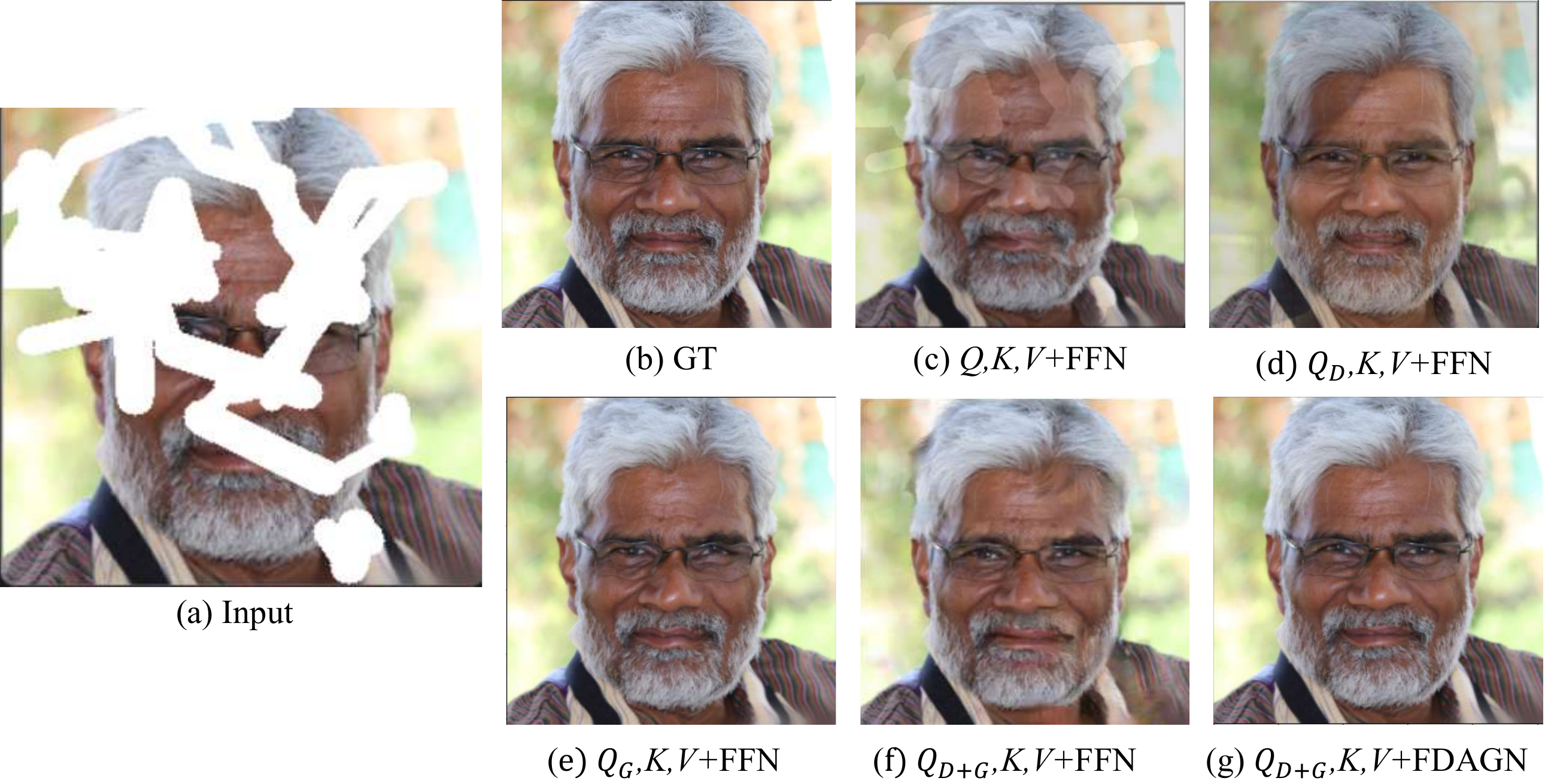}}
	\end{minipage}
		\caption{Ablation study with qualitative results on the CelebA dataset under {\rm{40\%}} to  {\rm{50\%}} occlusion ratio for image inpainting with different network configurations}
	\label{fig:res}
\end{figure}
\subsubsection{Loss function ablation experiment}
To verify the effectiveness of the proposed multi-loss integration strategy in image restoration, a set of ablation studies on loss combinations was conducted under consistent training settings. Specifically, perceptual loss, edge loss, and structural similarity loss were introduced incrementally to analyze their individual and combined contributions. The quantitative results are presented in Table IV, while the qualitative comparisons are shown in Fig. 9.

When using only L1 loss, the model achieved baseline performance, primarily focusing on pixel-level reconstruction. The inclusion of perceptual loss led to improvements in both PSNR and SSIM, indicating enhanced recovery of high-level semantic features. With the addition of edge loss, the model further improved in detail preservation, reflecting better boundary sharpness. Finally, incorporating structural similarity loss resulted in the highest overall performance, demonstrating enhanced structural coherence and texture fidelity. These results confirm the complementary benefits of each loss component in guiding more accurate and perceptually realistic image restoration.
\begin{table}[htbp]
	\centering
	\caption{Quantitative Results of Ablation Experiments on Different Loss Functions on the Rain200H Dataset}
	\renewcommand{\arraystretch}{1.3}
	\setlength{\tabcolsep}{8pt}
	\begin{tabularx}{\linewidth}{
			>{\centering\arraybackslash}X 
			>{\centering\arraybackslash}X 
			>{\centering\arraybackslash}X }
		\hline
		\multirow{2}{*}{Loss Configuration} & \multicolumn{2}{c}{Evaluation} \\
		\cline{2-3}
		& PSNR & SSIM \\
		\hline
		${L_1}$      & 31.97 & 0.9316 \\
		${L_1} + {L_P}$ & 32.20 & 0.9348 \\
		${L_1} + {L_P} + {L_E}$ & 32.28 & 0.9352 \\
		${L_1} + {L_P} + {L_E} + {L_M}$ & \textbf{32.34} & \textbf{0.9364} \\
		\hline
	\end{tabularx}
	\label{tab:addlabel}
\end{table}
\begin{figure}[htb]
	\begin{minipage}[b]{1.0\linewidth}
		\centering
		\centerline{\includegraphics[width=0.95\textwidth]{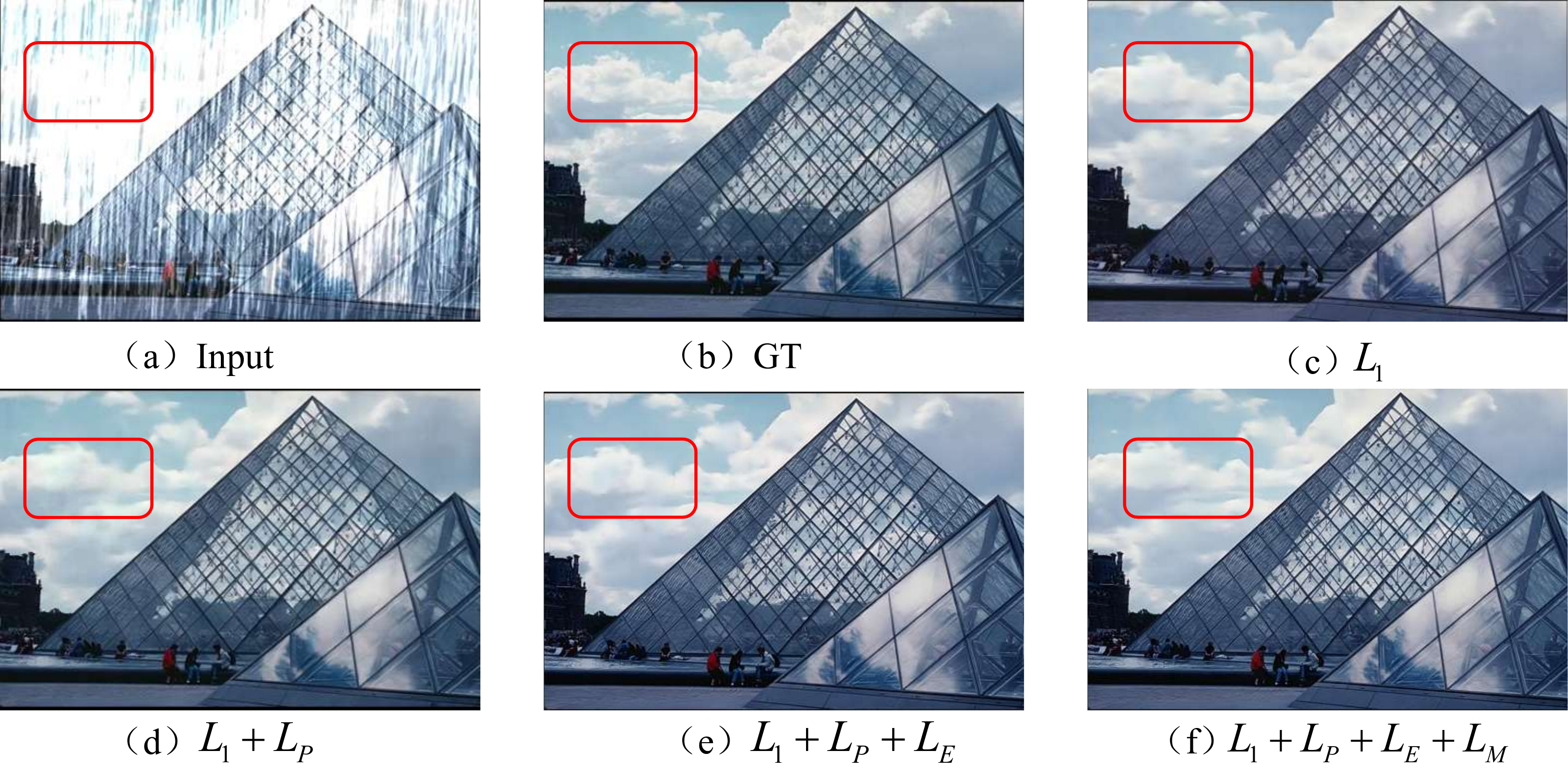}}
	\end{minipage}
	\caption{Qualitative Results of Ablation Experiments on Different Loss Functions on the Rain200H Dataset}
	\label{fig:res}
\end{figure}
\subsection{Parameter Sensitivity Analysis}
We propose a frequency-domain fusion approach for image inpainting, in which the high-frequency components of the input image are decomposed into multiple sub-bands via multi-scale wavelet transform. Fixed-wavelength Gabor filters are subsequently applied to each sub-band to extract directional texture features. To investigate the impact of key Gabor filter parameters within the proposed method, we conduct a series of sensitivity experiments by varying the wavelength and orientation. These experiments aim to evaluate how different parameter configurations affect the quality of texture extraction and the overall fidelity of image restoration.
\subsubsection{wavelength}
In this experiment, with fixed orientation and other parameters, only different fixed wavelengths of the Gabor filter were used to compare their effects on texture feature extraction and image reconstruction. To enhance model flexibility, a learnable wavelength parameter was introduced, allowing the model to adaptively adjust the scale configuration of the Gabor filter during training. This study aims to explore the impact of different filter configurations on texture detail restoration and overall visual quality in the same task scenario, providing references for parameter selection and optimization. The experimental results are shown in Table V.

The results show that as the fixed wavelength parameter increases, both PSNR and SSIM generally improve, indicating that larger wavelengths help enhance image restoration quality. Performance is relatively better at wavelengths of 1.5 and 2.0. However, the fixed wavelength approach has limitations: smaller wavelengths capture more high-frequency details but tend to amplify noise and artifacts, while larger wavelengths offer better smoothness and noise suppression but cannot flexibly adapt to regions with varying texture complexity, resulting in insufficient detail recovery. Therefore, fixed wavelengths struggle to balance detail preservation and noise suppression, limiting restoration effectiveness.

In contrast, the introduction of a learnable adaptive wavelength mechanism achieves the best PSNR and SSIM performance, outperforming all fixed wavelength settings. This mechanism dynamically adjusts the receptive field based on local image characteristics, favoring smaller wavelengths in texture-rich regions to enhance details and larger wavelengths in smooth areas to suppress noise, enabling fine-grained regional adaptation. The adaptive wavelength, driven by data, models texture features more precisely, avoiding over-enhancement or blurring caused by fixed scales, thus improving overall restoration quality. Moreover, the end-to-end optimization of wavelengths further strengthens the model’s robustness and adaptability in complex scenarios.
\begin{table}[htbp]
	\centering
	\caption{Performance of Different Parameter Configurations}
	\renewcommand{\arraystretch}{1.3}
	\setlength{\tabcolsep}{8pt}
	\begin{tabularx}{\linewidth}{
			>{\centering\arraybackslash}X
			>{\centering\arraybackslash}X
			>{\centering\arraybackslash}X}
		\hline
		Parameter Configuration & PSNR & SSIM \\
		\hline
		${\lambda}=0.5$  & 32.15 & 0.9349 \\
		${\lambda}=1.0$  & 32.13 & 0.9348 \\
		${\lambda}=1.5$  & 32.25 & 0.9353 \\
		${\lambda}=2.0$  & 32.28 & 0.9355 \\
		${\lambda}=2.5$  & 32.26 & 0.9353 \\
		Adaptive Wavelength & \textbf{32.34} & \textbf{0.9364} \\
		\hline
	\end{tabularx}
	\label{tab:param-config}
\end{table}

\subsubsection{orientation}
Based on wavelength analysis, we further investigated the sensitivity of directional parameters. The direction of a Gabor filter determines its ability to capture texture orientation and plays a key role in modeling high-frequency details. To assess how different direction settings affect image restoration, various configurations were applied to wavelet high-frequency subbands (LH, HL, HH), with all other model settings unchanged.

The tested strategies included: mismatched directions, unified direction, convolution-only replacement, multi-directional fusion (0°, 30°, 45°, 90°, 180°), random directions, and the proposed subband-specific optimal configuration. The results are shown in Table VI. Results show that mismatched directions significantly degrade performance, highlighting the importance of alignment between subband characteristics and filter orientation. Unified direction and convolutional alternatives provide moderate results but lack fine directional sensitivity. Multi-direction fusion offers richer responses but suffers from redundancy. Random directions lead to unstable learning and the worst outcomes.

The proposed “subband-direction” matching strategy applies vertical, horizontal  and diagonal Gabor filters to LH, HL, and HH subbands, respectively. This configuration achieves the best performance across all settings, enhancing texture reconstruction while maintaining model simplicity. These results validate the effectiveness of directionally aligned filtering based on wavelet subband properties.
\begin{table}[htbp]
	\centering
	\caption{Evaluation of Different Directional Strategies}
	\renewcommand{\arraystretch}{1.3}
	\setlength{\tabcolsep}{6pt}
	\begin{tabularx}{\linewidth}{
			>{\centering\arraybackslash}X
			>{\centering\arraybackslash}X
			>{\centering\arraybackslash}c
			>{\centering\arraybackslash}c}
		\hline
		Description & Strategy & PSNR & SSIM \\
		\hline
		Misaligned directions & LH→0°, HL→90°, HH→45° & 31.60 & 0.9279 \\
		& All bands→0° & 31.64 & 0.9282 \\
		Unified direction & All bands→90° & 31.59 & 0.9281 \\
		& All bands→45° & 31.66 & 0.9281 \\
		Conv-only baseline & All bands→Depth-wise Conv & 31.87 & 0.9306 \\
		Multi-direction fusion & LH/HL/HH→Fused directions & 32.24 & 0.9360 \\
		Random direction & LH/HL/HH→Randomly selected & 32.13 & 0.9347 \\
		Dabformer & LH→90°, HL→0°, HH→45° & \textbf{32.34} & \textbf{0.9364} \\
		\hline
	\end{tabularx}
	\label{tab:direction-strategy}
\end{table}

\section{CONCLUSION AND FUTURE DIRECTIONS}
In this paper, a frequency-domain fusion-based Transformer method is proposed for image inpainting. By integrating the multi-scale decomposition capability of wavelet transform with the directional selectivity of Gabor filters, the method effectively enhances high-frequency feature representation. Furthermore, a frequency-domain adaptive gating mechanism based on FFT is introduced to suppress noise and retain critical details, achieving superior structural restoration and texture reconstruction. Experimental results across multiple datasets validate the effectiveness of the proposed approach. Future work will explore learnable frequency-domain filters for improved adaptability, incorporate multimodal cues to enhance semantic understanding, and adopt more expressive loss functions and real-world degradation patterns to further improve model generalization and practical performance.

\bibliography{reference}

\vfill
\end{document}